\documentclass[twoside]{article}
\pdfoutput=1
\usepackage{amsmath}
\usepackage{mathtools}
\usepackage{cite}
\usepackage{natbib}
\usepackage[accepted]{aistats2015}
\begin{document}
%\linenumbers



\twocolumn[

\aistatstitle{Spatio-temporal Gaussian processes modeling of dynamical systems in systems biology }

\aistatsauthor{ Mu Niu \And Zhenwen Dai \And Neil Lawrence }

\aistatsaddress{ University of Sheffield \And University of Sheffield \And University of Sheffield} 
\aistatsauthor{ kolja becker  }

\aistatsaddress{ Institute of Molecular Biology, Mainz } ]

\begin{abstract}
Quantitative modeling of post-transcriptional regulation process is a challenging problem in systems biology. A mechanical model of the regulatory process needs to be able to describe the available spatio-temporal protein concentration and mRNA expression data and recover the continuous spatio-temporal fields. Rigorous methods are required to identify model parameters. A promising approach to deal with these difficulties is proposed using Gaussian process as a prior distribution over the latent function of protein concentration and mRNA expression. In this study, we consider a partial differential equation mechanical model with differential operators and latent function. Since the operators at stake are linear, the information from the physical model can be encoded into the kernel function. Hybrid Monte Carlo methods are employed to carry out Bayesian inference of the partial differential equation parameters and Gaussian process kernel parameters. The spatio-temporal field of protein concentration and mRNA expression are reconstructed without explicitly solving the partial differential equation.
\end{abstract}

\section{Introduction}

Quantitative modeling of post-transcriptional regulation is highly topical in systems biology.
Considering the vast possibilities of post-transcriptional gene regulation it is evident that protein expression patterns do not necessarily coincide with the location and timing of mRNA transcription. However, in many systems biology models the processes of transcription and translation have been considered together as a single step. This is partly due to the lack of relevant data but also owes to the higher complexity of the problem when allowing for multiple reactions.
The exact timing and position of gene expression is crucial in a developmental context and therefore post-transcriptional regulation of gene expression need to be taken into account. Here we review a recently published model \citep{becker13} of such regulation where mRNA and protein concentrations are explicitly considered separate state variables and data is available for both levels of description.

In \cite{becker13} the dynamical model is fitted to spatio-temporal expression data using a weighted least square estimate and the inference is limited to the discrete spatio-temporal grid points. Here we present an efficient Gaussian process based method for Bayesian inference of latent function and associated model parameters.
Gaussian processes have been effectively used in machine learning and statistical applications. \cite{lawrence06,graepel03} and \cite{murray05} are closely related to this work. \cite{lawrence06} have explored modeling a temporal dynamic system (ordinary differential equation) of transcriptional processes using Gaussian processes. An ordinary differential equation is represented by a integral operator operating on a latent function in \cite{lawrence06}. The model parameters are estimated using maximum likelihood optimization. In this study, we expand the idea of using Gaussian processes for dynamical system modeling to the spatio-temporal case. The partial differential equation model is seen as a linear differential operator applied to a latent function. Considering such linear operators simplifies the derivation of the covariance kernel. Modeling dynamical systems in systems biology with Gaussian processes lead to several advantages. First, it allows for the inference of continuous quantities without discretization which account for the spatial and temporal structure of the data. Second, the measurement error is naturally inherent to such models and third, it does not require an additional interpolation step to estimate the protein production rate. 

We describe in this paper a Gaussian process based approach for estimating the parameters of a model of post-transcriptional regulation described by a partial differential equation. In the second section, the model is rewritten in terms of linear operator and latent function. In the third section we show how linear operators can be included into Gaussian process regression to encode the biological information into the model. Since \cite{becker13} suggest the model parameters are correlated, Hybrid Monte Carlo (HMC) methods are used to carry out Bayesian inference of the partial differential equation parameters and Gaussian process hyper parameters in section four. The model is tested with simulated data and section five is dedicated to the application that originally motivated this work.

\section{Discriminative Models}

In the developing \emph{Drosophila} embryo the early positioning of body segments is partially controlled by the so called gap genes \citep{nuss80,jaeger11}. These gap genes are expressed in broad overlapping domains along the embryos anterio-posterior axis with the precise positioning resulting from regulation by upstream maternal transcription factors but also from cross-regulation between the different gap genes. This gene regulatory network involved in early \emph{Drosophila} development has been extensively studied (for example in \cite{jaeger04,jaeger07,surk09}). However, these studies consider either the expression domains of gap gene mRNA or the gap gene proteins separately due to the lack of relevant expression data and the increasing complexity of the problem. More recently, \cite{becker13} presented a model in which both layers of description (genes ans proteins) are resolved simultaneously and post-transcriptional regulation is taken
into account. The expression of mRNA and protein along the embryos main axes can be quantified in detail using fluoresesent microscopy \citep{flyex1,flyex2,becker13}. In \cite{becker13}, protein production of a single gap gene is considered to be linearly dependent on its input mRNA concentration at an earlier time point. The model also allows for diffusion of protein between cells and linear protein decay. These processes are dependent on the diffusion parameter and the degradation rate of protein respectively. In order to reformulate the model in the context of Gaussian processes a rescaling of model parameters was carried out:
\begin{equation}\label{pde}
 -D \frac{\partial ^2 y(x,t)}{\partial x^2} + \alpha \frac{\partial y(x,t)}{\partial t} + \beta y(x,t)=  f(x,t) 
\end{equation}

where $f(x,t)$ is mRNA concentration. $y(x,t)$ is protein concentration. They are both function of time and space. $D$ denotes to the diffusion rate of protein (corresponding to $D/\alpha$ in \cite{becker13}). $\alpha$ is the inverse of the rate of protein production and $\beta$ is the scaled protein degradation rate. Considering Fick's laws of diffusion, the diffusion rate $D$ is fixed to be positive. In the process of translation protein molecules are produced from an mRNA template. The production rate can therefore at minimum be zero corresponding to no protein being produced. Therefore, $\alpha$ and $\beta$ need to be positive.

Equation \ref{pde} can be rewritten as 
\begin{equation}
f(x,t)=\mathcal L_{x,t} y(x,t)
\end{equation}

where $\mathcal L_{x,t} $ is a linear operator that returns the function $f(x,t)$ when applied to the function $y(x,t)$. In general $\mathcal L_{x,t} $ can be any kind of linear operator such as integral operator
 $\mathcal L_{x} y(x) = \int_{-\infty}^{x} y(x_i)dx_i $ or differential operator $\mathcal L_{x} y(x)=\partial y(x)/ \partial x$ \citep{sarkka11}. In our case, $\mathcal L_{x}$ is a sum of partial differential operator defined as 
\begin{equation}
\mathcal L_{x,t} = -D\frac{ \partial^2}{\partial x^2} + \alpha\frac{ \partial}{\partial t} + \beta
\end{equation}

\section{Linear Operator and Gaussian Process}
In this article, we will model the protein concentration $y(x,t)$ as a latent function drawn from a zero mean Gaussian process prior distribution with covariance function $E[y(x,t)y^T(x',t')]=k_{yy}(x,t,x',t')$, which is denoted as 
\begin{equation}
y(x,t) \sim \mathcal{GP}(0,k_{yy}(x,t,x',t'))
\end{equation}

$k_{yy}(x,t,x',t')$ is defined as a $2$ dimensional separable RBF (radial basis function, also known as exponentiated quadratic or squared exponential) spatio-temporal kernel, which is described as

\begin{equation}\label{k_yy}
\begin{split}
k_{yy}(x,t,x',t') = \sigma_{y}^2 exp\Big(- \frac{(t-t')^2}{2\theta_t^2} - \frac{(x-x')^2}{2\theta_{x}^2 }  \Big)
%k_{yy}(x,t,x',t') =  k_{yy_t}(t,t') \otimes k_{yy_x}(x,x') \\
%=\sigma_{y}^2 exp\Big(- \frac{(t-t')^2}{2\theta_t^2} \Big)    \otimes  exp\Big(- \frac{(x-x')^2}{2\theta_{x}^2 }  \Big)
\end{split}
\end{equation}

It is the tensor product of two separate RBF kernels  with different lengthscales $\theta_{t}^2$ and $\theta_{x}^2$  for the temporal and spatial directions. 
Applying the rules of linear transformation of Gaussian processes \citep{rasmussen06, papoulis02} to $f(x,t)=\mathcal L_{x,t} y(x,t)$ gives
\begin{multline}
k_{ff}(x,t,x',t') = cov(f(x,t),f(x',t')) \\
= cov( \mathcal L_{x,t}y(x,t) , \mathcal L_{x',t'}y(x',t')  ) \\
=\mathcal L_{x,t}  \mathcal L_{x',t'} k_{yy}(x,t,x',t')
\end{multline}

Here $\mathcal L_{x',t'}$ only operates on the second pair of arguments of the kernel $k_{yy}(x,t,x',t')$.
Explicitly, $k_{ff}(x,t,x',t')$ is given by the following formula
\begin{equation}\label{k_ff}
\begin{split}
k_{ff}(x,t,x',t') = \Big( -D\frac{ \partial^2}{\partial x^2} + \alpha\frac{ \partial}{\partial t} + \beta \Big) \\
\Big( -D\frac{ \partial^2}{\partial x'^2} + \alpha\frac{ \partial}{\partial t'} + \beta \Big) k_{yy}(x,t,x',t')
\end{split}
\end{equation}

As it is shown in equation \ref{k_ff}, the covariance kernel of $f(x,t,x',t')$ is the result of applying the partial differential operator twice on the kernel of $y(x,t,x',t')$. The analytical result writes:
\begin{multline} \label{k_ffa}
%\begin{split}
k_{ff}(x,t,x',t') = \sigma_y^2 \left( \alpha^2 \Big(\frac{1}{\theta_t^2} - \frac{(t-t')^2}{\theta_t^4} \Big)- \right.\\
  2D \beta \Big(\frac{(x-x')^2}{\theta_x^4}- \frac{1}{\theta_x^2} \Big) +  \\
\left. D^2 \Big( \frac{3}{\theta_x^4} - \frac{6(x-x')^2}{\theta_x^4} + \frac{(x-x')^4}{\theta_x^4} \Big) + \beta^2 \right)  \\  
 \exp \Big( -\frac{(t-t')^2}{2\theta_t^2} - \frac{(x-x')^2}{2\theta_x^2} \Big)
%\end{split}
\end{multline}

To infer the mRNA expression function $f$, the ``cross-covariance'' term between $y$ and $f$ can be derived as
\begin{equation} \label{k_fy}
k_{fy}(x,t,x',t') = \mathcal L_{x,t}  k_{yy}(x,t,x',t')
\end{equation}
\begin{equation}\label{k_yf}
k_{yf}(x,t,x',t') = \mathcal L_{x',t'}  k_{yy}(x,t,x',t')
\end{equation}

Again, the cross covariance function $k_{yf}$ can be obtained 
explicitly for the RBF prior on the latent function $y$: 
\begin{multline} \label{k_yfa}
k_{yf}(x,t,x',t') = \Big (  -D(\frac{(x-x')^2}{\theta_x^4} - \frac{1}{\theta_x^2})\\  + \alpha \frac{(t-t')}{\theta_t^2}  + \beta  \Big)\\ \sigma_y^2 exp\Big(-\frac{(t-t')^2}{2\theta_t^2} - \frac{(x-x')^2}{2\theta_x^2} \Big)
\end{multline}

By combining $f$ and 
$y$ as a state vector $s$, a multi-output Gaussian process model can be constructed as per equation \ref{kernelM}. The variance function $k_{yy}$ and $k_{ff}$ can be calculated according to equation \ref{k_yy} and \ref{k_ffa}. The cross covariance function $k_{yf}$ and $k_{fy}$ are derived using equation \ref{k_yfa}. 
\begin{equation} \label{kernelM}
s(x,t)=\begin{bmatrix}
 y(x,t)\\
 f(x,t)
\end{bmatrix}, \
\mathcal K =\begin{bmatrix}
k_{yy} & k_{yf} \\
k_{fy} & k_{ff}
\end{bmatrix} 
\end{equation}
\begin{equation}
 s(x,t) \sim \mathcal GP ( 0, \mathcal K(x,t,x',t'))
\end{equation}

where $\mathcal K$ is  the matrix of covariance function.

\subsection{Gaussian Process regression}
The latent regression function $f$ and $y$ have been modeled as zero mean multi-output Gaussian process. The multi-output Gaussian process regression model can be stated as
\begin{equation} \label{likelihood}
\begin{split}
s(x,t)\sim \mathcal{GP}(0,\mathcal K(x,t,x',t')) \\
 S(x_i,t_i) = s(x_i,t_i) + \mathcal E_i \\
 \mathcal E_{1:n\times m} \sim \mathcal N(0,\Sigma)
\end{split}
\end{equation}

Let us assume that we have noisy observations $Y$ and $F$ on a grid of $n$ in spatial direction and $m$ in temporal direction. $S$ is vector representation of the observation. The measurement error $\mathcal E= (\mathcal E_1,...,\mathcal E_{n\times m})$ is given by the covariance matrix $\Sigma$. In our case, $\Sigma$ is a diagonal matrix.  Given the vector of measurements $S= (Y_1, ..., Y_{n\times m},F_1, ... ,F_{n\times m})$, the posterior mean and variance are given by the Gaussian process regression equations \citep{o78,rasmussen06}:
\begin{multline*}
E[s(x,t)|S] = \mathcal K(x,t,x'_{1:n\times m},t'_{1:n\times m}) \\
 [\mathcal K(x'_{1:n\times m},t'_{1:n\times m},x'_{1:n\times m},t'_{1:n\times m}) + \Sigma]^{-1} S 
\end{multline*}
\begin{multline} 
Var[s(x,t)|S] =  \mathcal K(x,t,x,t) -  \\
 \mathcal K(x,t,x'_{1:n\times m},t'_{1:n\times m}) \\ [\mathcal K(x'_{1:n\times m},t'_{1:n\times m},x'_{1:n\times m},t'_{1:n\times m}) + \Sigma]^{-1} \\ \mathcal K^{T} 
 (x,t,x'_{1:n\times m},t'_{1:n\times m}) 
\end{multline}

Although the main objective of this article is to estimate the partial differential equation parameters, a similar method can also be applied to give the probabilistic solution of the partial differential equation given the known model parameters. 
For example if we only have measurements of $f$ and we want to predict $y$ given the model parameters. Without solving the partial differential equation, we can use Gaussian Process regression model to predict $y$. The conditional mean and covariance of $y$ become:
\begin{multline*}
E[y(x,t)|F] =  k_{yf}(x,t,x'_{1:n\times m},t'_{1:n\times m}) \\
 [k_{ff}(x'_{1:n\times m},t'_{1:n\times m},x'_{1:n\times m},t'_{1:n\times m}) + \Sigma]^{-1} F
\end{multline*}
\begin{multline}
Var[y(x,t)|F] =  k_{yy}(x,t,x,t) -  \\
 k_{yf}(x,t,x'_{1:n\times m},t'_{1:n\times m})\\
   [ k_{ff}(x'_{1:n\times m},t'_{1:n\times m},x'_{1:n\times m},t'_{1:n\times m}) + \Sigma]^{-1} \\ 
  k_{fy}^{T} (x,t,x'_{1:n\times m},t'_{1:n\times m}) 
\end{multline}

\section{Hybrid Monte Carlo}

High correlations between model parameters have been suggested by \cite{becker13}. The normal Markov Chain Monte Carlo sampling takes very long time to traverse the parameter space. Increasing the proposal variance to make bigger transitions may also result in low rates of acceptance and poor mixing of the chain. The consequence of the inefficient random walk proposal is a small effective sample size from the chain  \citep{liu08,robert04}. Hybrid Monte Carlo is believed to be more efficient by producing distant proposal, thereby avoiding the slow exploration of the parameter space \citep{neal11,neal96}.

In Hybrid Monte Carlo, the deterministic proposal based on Hamiltonian equation is applied along with stochastic proposal to provide ergodic Markov chain \citep{duane87}. The intuition is that the density of target distribution is treated as the potential energy and auxiliary random variables are introduced of which the density is treated as the kinetic energy. Each parameter $\theta_i$ (we use $\theta$ to represent all the model parameters and kernel hyper parameters) is paired with a momentum variable $ h_i$. The total energy is defined as the negative joint log probability:
\begin{equation}
H(\theta,h) = -\mathcal L(\theta) + \frac{1}{2}log(2\pi)^d |M| + \frac{1}{2}h^TM^{-1}h
\end{equation}

where $\mathcal L(\theta)$ is the log of target density. $h^TM^{-1}h$ is the kinetic energy term. The covariance matrix $M$ denotes a mass matrix and $d$ is the dimension of the parameter $\theta$. The physical analogy of this negative joint log-probability is a Hamiltonian \citep{duane87, leimkuhler04}. The time evolution of the system is defined by the Hamilton equations \citep{neal11}
\begin{equation}
\begin{split}
\frac{d\theta}{d\tau}=\frac{\partial H}{\partial h} = M^{-1}h \\
\frac{dh}{d\tau}=-\frac{\partial H}{\partial \theta}=  \nabla_{\theta} \mathcal L(\theta(\tau))
\end{split}
\end{equation}

where $\nabla_{\theta} \mathcal L(\theta(\tau))$ is the derivative of $\mathcal L(\theta(\tau))$ evaluated at $\theta(\tau)$.
If the parameters $\theta$ move along the paths based on equation above, they will essentially move along the contours of the target distribution. In practice the Hamiltonian equation is solved numerically to propose movement for $\theta$. A few steps of parameter updating, known as leapfrog steps, are carried on, in which the auxiliary variable $h$ and the parameters $\theta$ are updated alternately.  
A leapfrog step is defined as:
\begin{equation}
\begin{split}
h(\tau +\frac{\epsilon}{2}) =h(\tau) + \frac{\epsilon}{2}\nabla_{\theta} \mathcal L(\theta(\tau)) \\
\theta(\tau + \epsilon)= \theta(\tau) + \epsilon M^{-1} h (\tau + \frac{\epsilon}{2})\\
h(\tau+\epsilon) = h(\tau + \frac{\epsilon}{2}) + \frac{\epsilon}{2}\nabla_{\theta} \mathcal L(\theta(\tau+\epsilon))
\end{split}
\end{equation}

where $\epsilon$ is the step size of Hamiltonian move step. After a given number of leapfrog steps, a metropolis update is performed, in which the proposed $\theta$ and $h$ is accepted based on the previous $\theta^*$ and $h^*$ with probability
\begin{equation*}
\min \Big( 1, \exp \big(-H(\theta, h) + H(\theta^*,h^*)\big) \Big)
\end{equation*}

 If the proposed state is not accept, the next state is kept the same as the previous one.  The step size $\epsilon$ and number of integration steps can be tuned based on the acceptance rate. Heuristics suggest that the choice of matrix $M$ should rely on the knowledge of marginal variance of the target distribution \citep{neal11,liu08,girolami11}. %The prior variance is used to build $M$ in our case.

As it can be seen in the Hamiltonian equation, the updating of $h$ needs the gradient of the log target density $\nabla_{\theta} \mathcal L(\theta)$ which is the gradient of log posterior of $\theta$.  This requires calculating the derivative of the log likelihood respect to each parameters. The log posterior is calculated based on equation \ref{post}.
%\begin{equation*}
 %S(x,t) \sim \mathcal{GP}(0,\mathcal K(x,t,x',t') + \Sigma)
%\end{equation*}
\begin{multline}\label{post}
\ln(\mathcal L(\theta)) = \frac{1}{2} \ln((2\pi)^{n\times m} |\mathcal K+ \Sigma|) +\\
 \frac{1}{2} S^T (\mathcal K+\Sigma)^{-1}S - \ln(p(\theta))
\end{multline}

where $p(\theta)$ is the prior.
The gradient of log likelihood is given by
\begin{multline}
\frac{\partial Lik}{\partial \theta} =  S^T \mathcal K^{-1} \frac{\partial \mathcal K}{\partial \theta} \mathcal K^{-1} S 
 = \sum \mathcal K^{-1} SS^{T} \mathcal K^{-1} \circ \frac{\partial \mathcal K}{\partial \theta} \\
 =  \sum \frac{\partial L}{\partial \mathcal K} \circ \frac{\partial \mathcal K}{\partial \theta}
\end{multline}

If we replace $ \mathcal K^{-1} SS^{T} \mathcal K^{-1}$ as $ \partial L/\partial \mathcal K$, the gradient of log likelihood become the element sum of hadamard product of  $\partial L/ \partial \mathcal K$ and $\partial \mathcal K/ \partial \theta$.  $\partial \mathcal K /\partial \theta$ is defined as

\begin{equation}
\frac{\partial \mathcal K}{\partial \theta}  =
\begin{bmatrix}
\frac{\partial k_{yy}}{\partial \theta} & \frac{\partial k_{yf}}{\partial \theta} \\
\frac{\partial k_{fy}}{\partial \theta} & \frac{\partial k_{ff}}{\partial \theta}
\end{bmatrix} 
\end{equation}

 $\partial \mathcal K /\partial \theta$ is the gradient of kernel. It is derivative of equation (\ref{kernelM}) which is the matrix of gradient of $k_{yy}$ , $k_{ff}$ and cross covariance.

\subsection{Implementation with simulation data}
Before tackling real data in section \ref{real}, we look at inference for simulated data. We consider trigonometric functions for $y$ and $f$ with some observation noise. $D$, $\alpha$ and $\beta$ are chosen to be $1$. We randomly pick some time points and spatial location. Noisy observations of $y$ and $f$ are generated using equation \ref{sime}.
\begin{equation}\label{sime}
\begin{split}
Y(x,t)= \cos (x) + \sin (t) + e \\
F(x,t)= \sin (t) + cos(t) +2\cos(x) +e \\
e \sim N(0,\sigma_0)
\end{split}
\end{equation}

The model parameters are estimated using HMC algorithm. We run it for 7000 iterations after burn-in. The trace plots of HMC are shown in Figure 3. The posterior mean and standard deviation of model parameters are shown in Table~\ref{sim-table}, along with the actual values used in simulation. Posterior density plots for the parameters are given in Figure 2. All the posterior distributions are consistent with the true values. The significant correlation between $\alpha$ and $\beta$ (Figure 2 d) implies that HMC is more effective than MCMC for these settings. 

\begin{figure}[h]\label{opsim}
%\vspace{.3in}
%\centerline{\fbox{This figure intentionally left non-blank}}
%\centerline { \includegraphics[width=0.5\textwidth]{./pic/simoall2.pdf} }
\centerline { \includegraphics[width=0.5\textwidth]{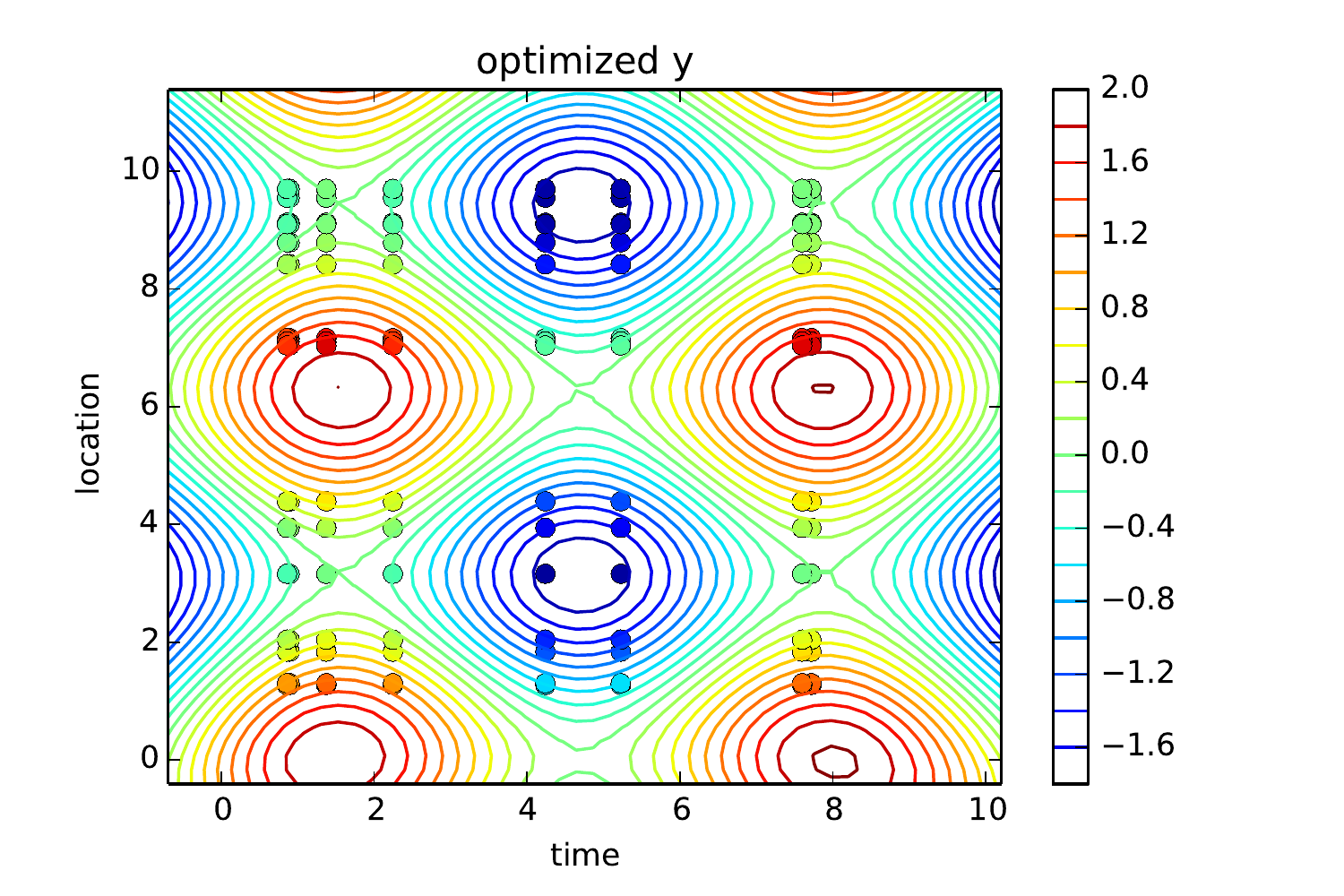} }
\centerline { \includegraphics[width=0.5\textwidth]{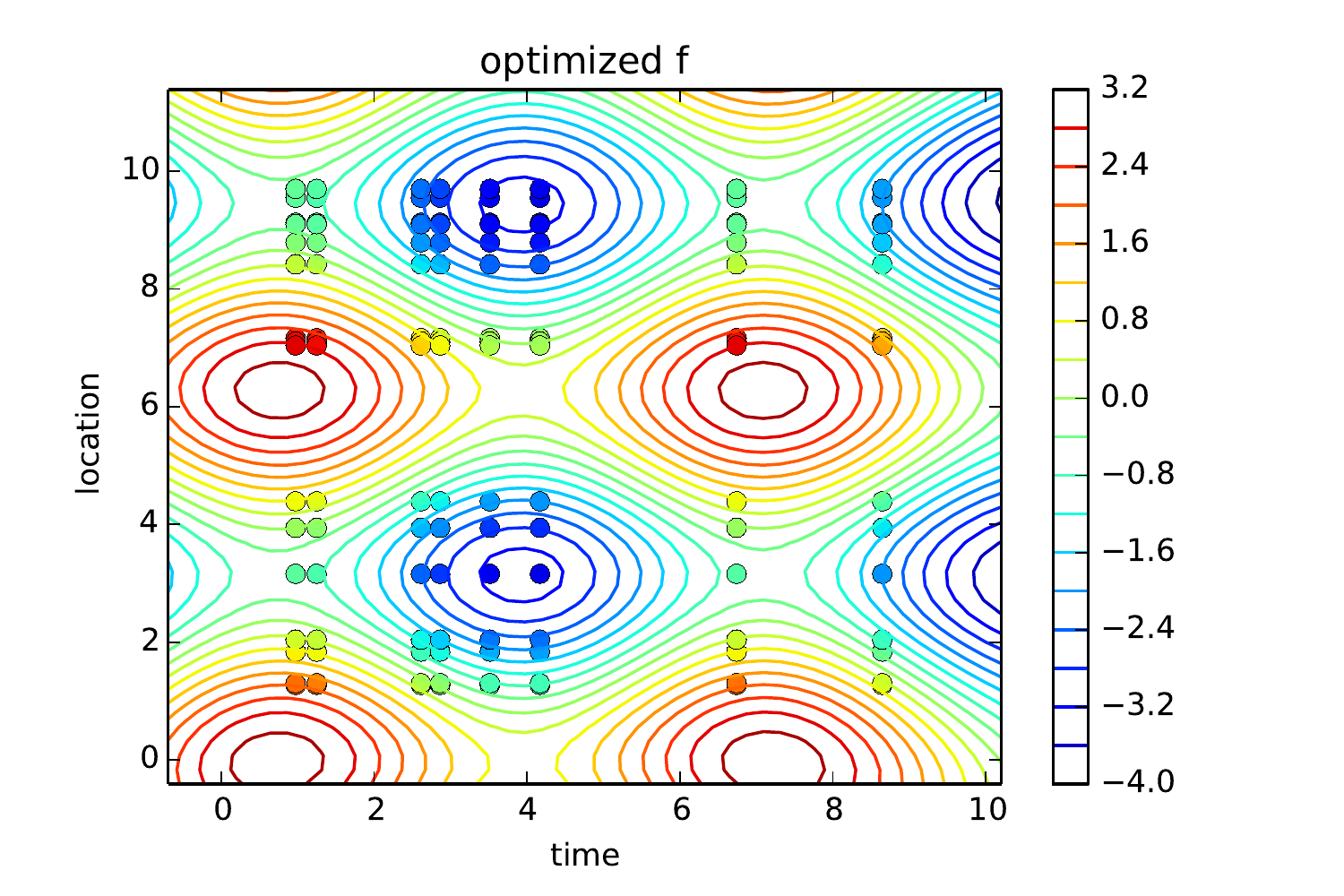} }
%\vspace{.3in}
\caption{spatio-temporal field of $y$ and $f$ with optimized model parameters for simulation data.}
\end{figure}

\begin{table}[h]
\caption{Model parameter estimate with simulation data} \label{sim-table}
\begin{center}
\begin{tabular}{cccc}
{\small \bf Parameter}  &{\small \bf Mean} & {\small \bf Standard deviation} &{\small \bf True}\\
\hline \\
D &0.993&0.02&1\\
$\alpha$  &1.002&0.028&1\\
$\beta$  &1.002&0.013&1\\
\end{tabular}
\end{center}
\end{table}

\begin{figure}[h]
%\vspace{.3in}
%\centerline{\fbox{This figure intentionally left non-blank}}
\centerline { \includegraphics[width=0.5\textwidth]{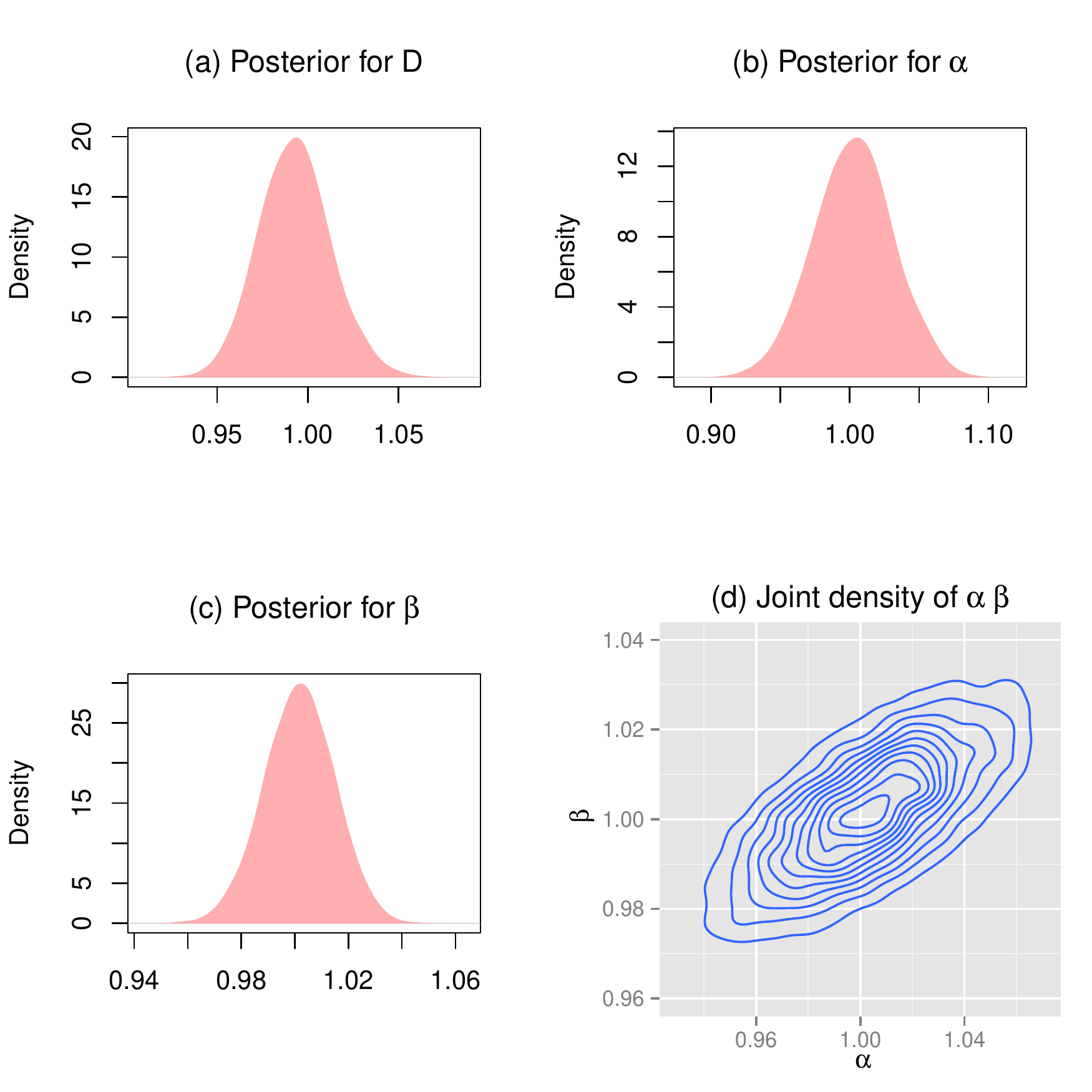} }
%\vspace{.3in}
\caption{Posterior densities for model parameters and joint posterior of $\alpha$ and $\beta$ with simulation data, based on the Hybrid Monte Carlo runs of 7000 iterations. (a) $D$ is the protein diffusion rate. (b) $\alpha$ is the inverse of protein production rate. (c) $\beta$ is the scaled protein degradation rate. (d) joint density of $\alpha$ and $\beta$.  } 
\end{figure}\label{simdensity}

\begin{figure}[h]\label{tracesim}
%\vspace{.3in}
%\centerline{\fbox{This figure intentionally left non-blank}}
\centerline { \includegraphics[width=0.4\textwidth]{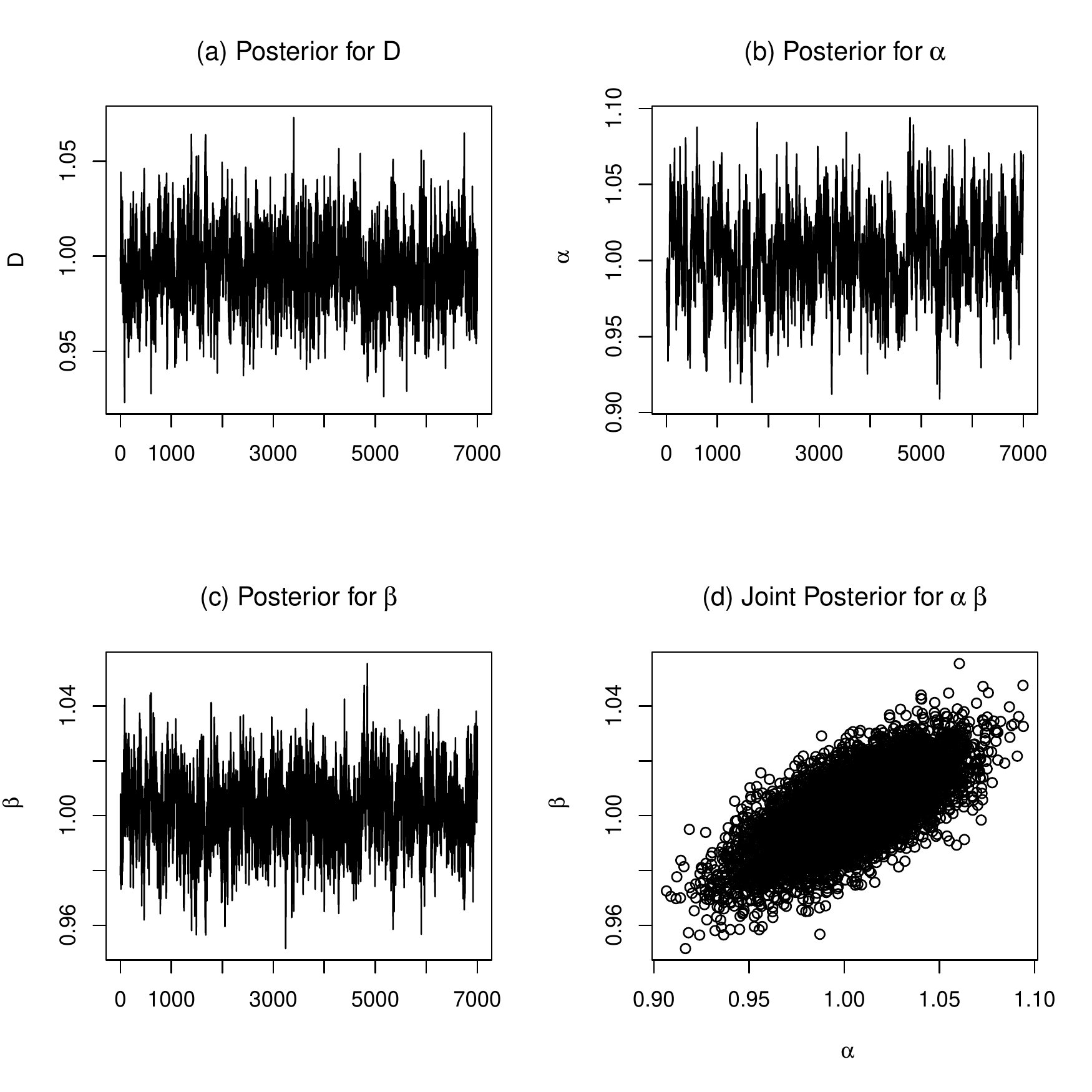} }
%\vspace{.3in}
\caption{Trace plot of model parameter for simulated data based on the Hybrid Monte Carlo runs of 7000 iterations. (a) $D$ is the protein diffusion rate. (b) $\alpha$ is the inverse of protein production rate. (c) $\beta$ is the scaled protein degradation rate. (d) joint posterior of $\alpha$ and $\beta$.} 
\end{figure}

The optimized spatio-temporal field is plotted in Figure 1 by setting model parameter as the posterior mean.
The colored contour in the figure represent the predicted mean of $y$ and $f$. The colored circle represent the data points simulated from trigonometric functions. 
These results show that the multioutput Gaussian process model and Bayesian inference capture the key mechanism at work here.

%\newpage

\section{Implementation with real data} \label{real}

In this section we have fitted the spatio-temporal multioutput Gaussian processes to existing data from \cite{becker13}. For this we have chosen the mRNA and protein expression data of the gap gene \emph{Knirps}.

Considering the available information for the measurement error, the heteroscedastic Gaussian process regression model is applied by assigning fixed variance (measurement error) for each data point. The prior distribution of model parameters are given based on the results in \cite{becker13}. 

The results given here are based on 10000 iterations of HMC runs after burn-in. The trace plots of HMC are shown in Figure 6. Table \ref{sample-table} shows posterior means and posterior standard deviations for the parameters of the partial differential equation. The weighted least square estimated mean of model parameters from \cite{becker13} is presented in the last column of Table~\ref{sample-table}.

\begin{figure}[h]\label{real_sp}
\centerline { \includegraphics[width=0.5\textwidth]{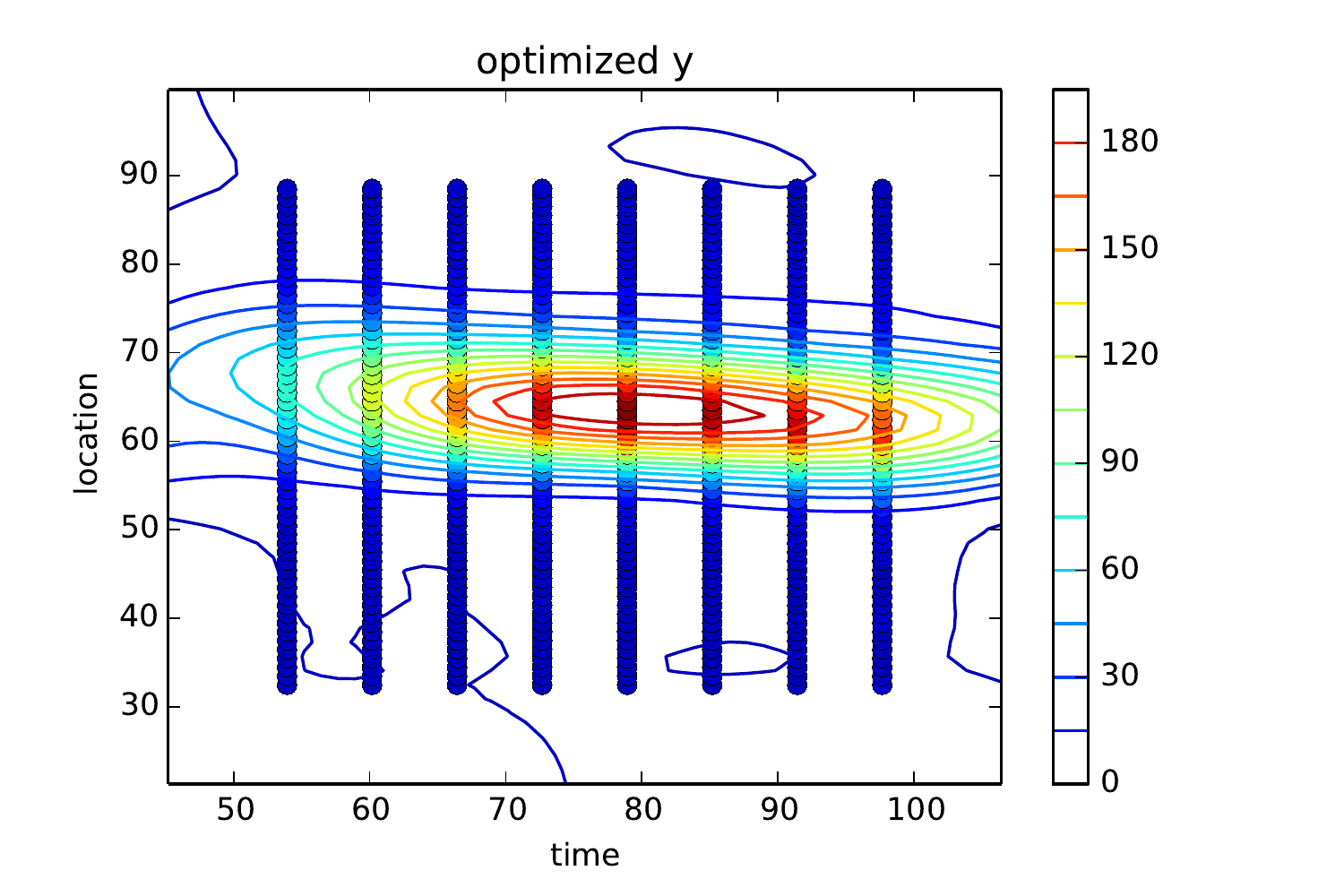} }
\centerline { \includegraphics[width=0.5\textwidth]{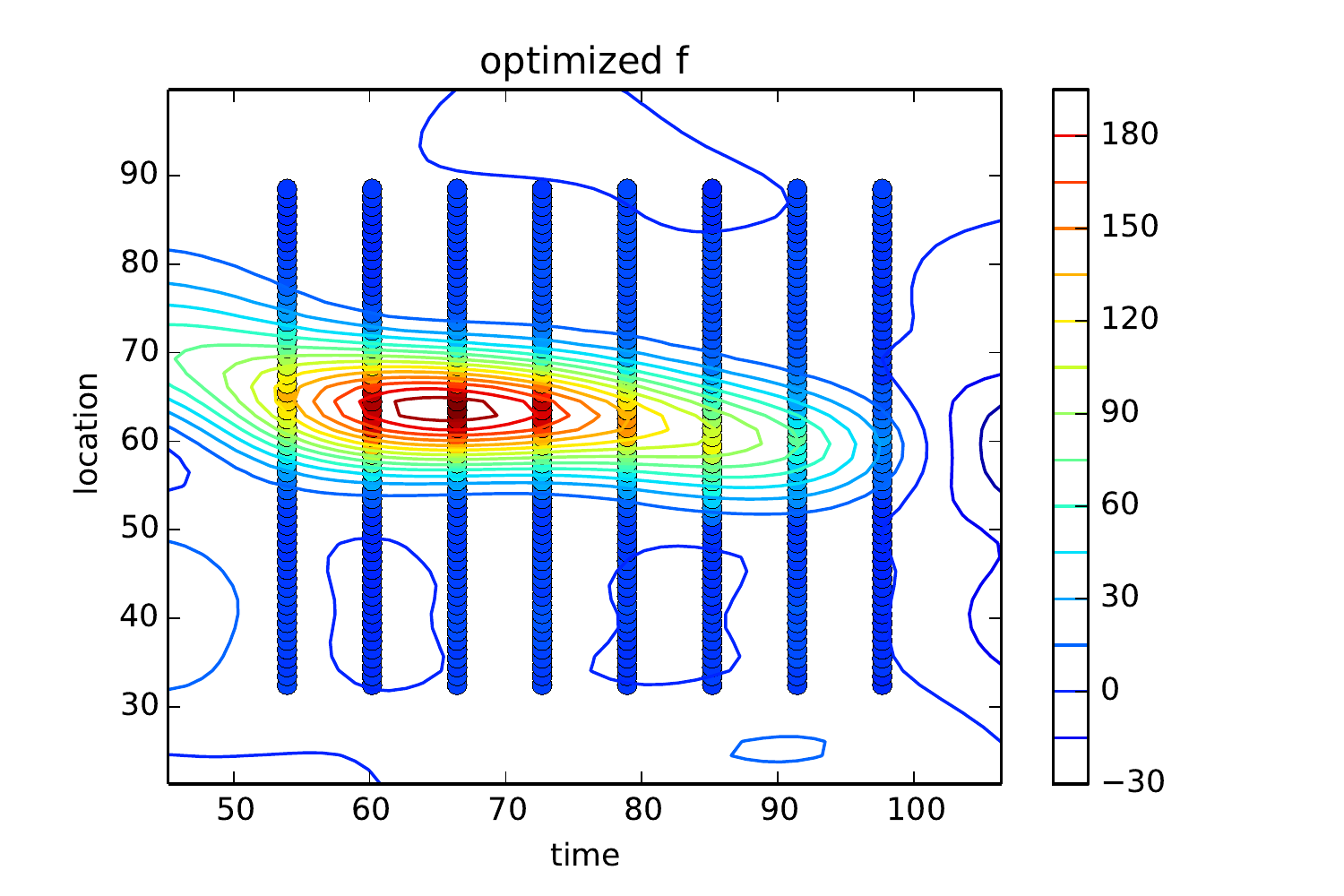} }
%\vspace{.3in}
\caption{spatio-temporal field of $y$ and $f$ with optimized model parameters for real data.}
\end{figure}

\begin{table}[h]
\caption{Model Parameter estimate with real data} \label{sample-table}
\begin{center}
\begin{tabular}{cccc}
{\bf Parameter}  &{\bf Mean} & {\bf Sd}&{\bf Becker[2013]}\\
\hline \\
D &0.017&0.023 &0.16\\
$\alpha$  &13.04&0.72&12.771 \\
$\beta$  &0.65&0.022&0.983\\
\end{tabular}
\end{center}
\end{table}

The density of model parameters are plotted in Figure 5. As can be seen in panel (d), the parameters $\alpha$ and $\beta$ are correlated. The spatio-temporal field of $y$ and $f$ are presented in Figure 4. As we have noised observations, the colored circle represent the mean of data points.

\begin{figure}[h]\label{realdensity}
\centerline { \includegraphics[width=0.5\textwidth]{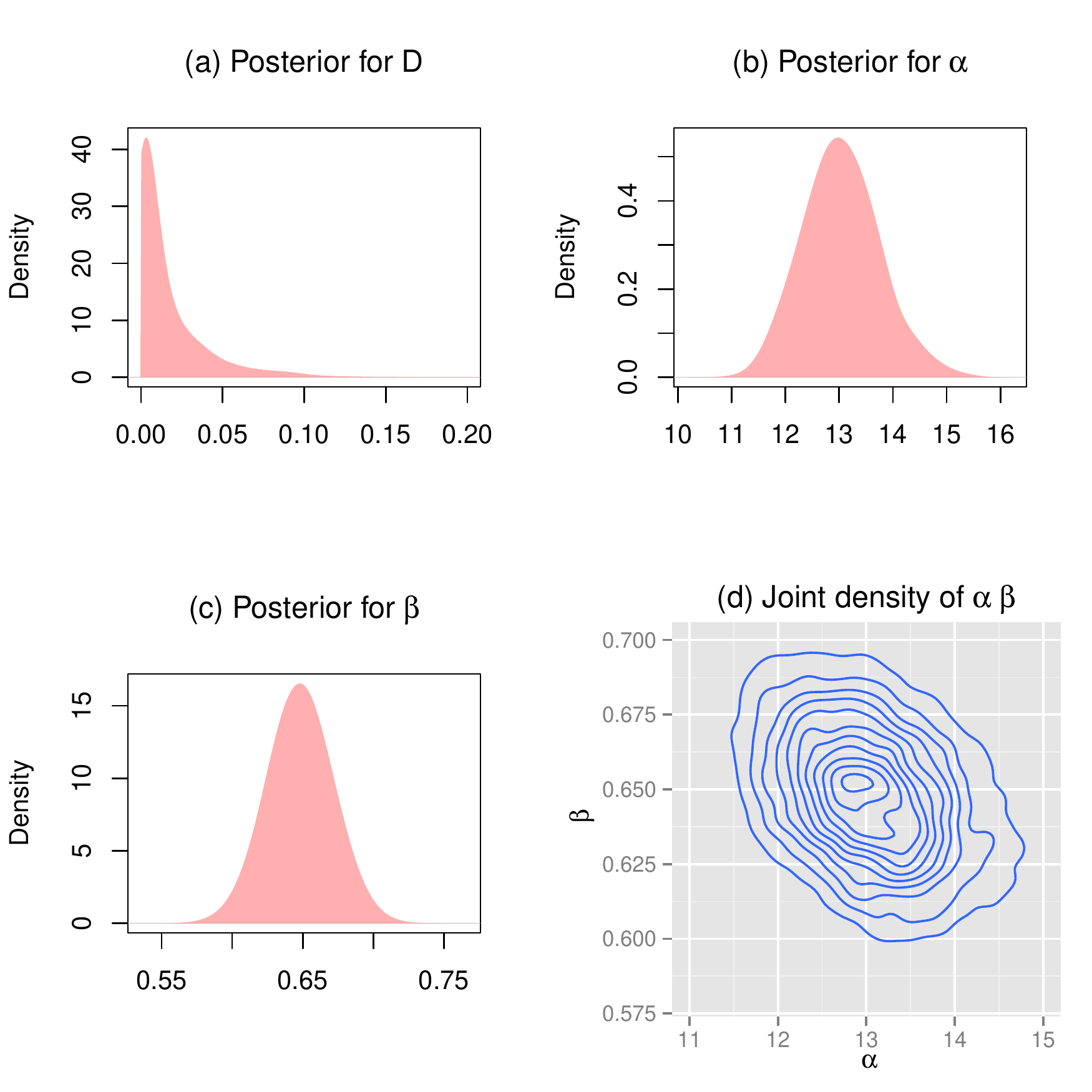} }
%\vspace{.3in}
\caption{Posterior densities for model parameters and joint posterior of $\alpha$ and $\beta$ with real data, based on the Hybrid Monte Carlo runs of 10000 iterations. (a) $D$ is the protein diffusion rate (b) $\alpha$ is the inverse of protein production rate (c) $\beta$ is the scaled protein degradation rate (d) joint density of $\alpha$ and $\beta$.}\label{density} 
\end{figure}

\begin{figure}[h]\label{realdensity}
\centerline { \includegraphics[width=0.4\textwidth]{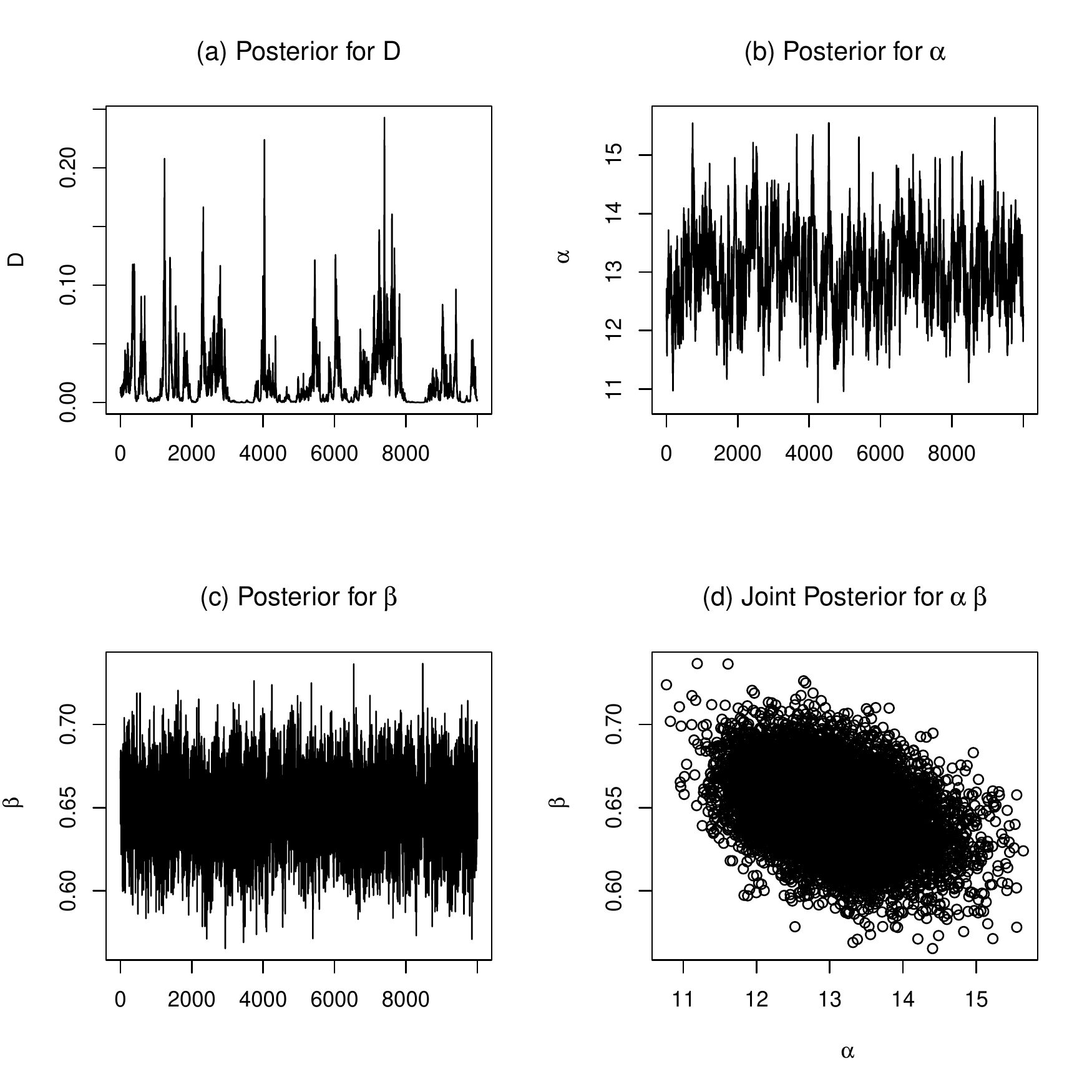} }
%\vspace{.3in}
\caption{Trace plot of model parameter for real data based on the Hybrid Monte Carlo runs of 10000 iterations. (a) $D$ is the protein diffusion rate. (b) $\alpha$ is the inverse of protein production rate. (c) $\beta$ is the scaled protein degradation rate. (d) joint posterior of $\alpha$ and $\beta$. }\label{density} 
\end{figure}

The order of magnitude of the estimated parameters $\alpha$ and $\beta$ are comparable to the estimated parameter values in ~\cite{becker13}.
%(Please note the rescaling of the parameters). 
According to the model, these parameters are confidently estimated since the standard deviation of their posterior is small.
On the other hand, there is a ten fold difference between the estimated parameter value $D=0.017$ in this article and the one in  \cite{becker13} ($D=0.16$ after rescaling). The difference might be caused by the different modeling approach. The exclusion of the delay parameter from the original model could also potentially account for the difference.
% But so far this reasoning is only speculative. 
However, the estimated parameter value for $D$ of the results presented here is well contained within the confidence interval given by \cite{becker13}. Also the significant correlation between production and decay rates has been correctly captured using the Gaussian process approach (see Figure 4 (d)).
Before concluding, we would like to stress that both the approach based on Gaussian processes presented in this paper and the parameter estimation and identifiability analysis method in \cite{becker13} reach qualitatively similar results. The biological interpretation regarding the time scale of protein decay and the practical non-existence of diffusion in the system therefore remain untouched.

\section {Conclusion}
In this article we discussed how Gaussian processes can be effectively used to model a spatio-temporal dynamical model of post-transcriptional regulation. Comparing with the standard parametric method, our nonparametric approach does not require restricting the inference to the observed spatio-temporal data points or any form of grid points. The continuity of the spatio-temporal field is naturally accounted for in our Gaussian process model. Hybrid Monte Carlo is employed to infer model parameters. The correlation between partial differential equation parameters are successfully captured from the posterior density.

Although the transcriptional delay has been dropped from the original partial differential equation, the delay operator is linear so it could be included in the analysis without requiring any additional theoretical results. 
%One of the future research would be including this delay parameter in the model. These effects could be naturally included in Gaussian process model by introducing the delay parameter into the distance function of the kernel. 
However, \cite{lawrence06} states the data need to be sampled at a reasonably high frequency to identify the delays. 
A promising development for future research is to introduce nonlinearities into the dynamic model as in the Michaelis-Menten model \citep{rogers06}.

\subsubsection*{Acknowledgements}
%We gratefully acknowledge the support we received from European Commission grant FP7-KBBE-2011-5/289434 ``Bioinformatics Methods and Tools for Data-Driven, Predictive Dynamic Modelling in Biotechnological Applications''.

\bibliography{GPPaper}{}

\begin{thebibliography}{23}
\providecommand{\natexlab}[1]{#1}
\providecommand{\url}[1]{\texttt{#1}}
\expandafter\ifx\csname urlstyle\endcsname\relax
  \providecommand{\doi}[1]{doi: #1}\else
  \providecommand{\doi}{doi: \begingroup \urlstyle{rm}\Url}\fi

\bibitem[Becker et~al.(2013)Becker, Balsa-Canto, Cicin-Sain, Hoermann,
  Janssens, Banga, and Jaeger]{becker13}
Kolja Becker, Eva Balsa-Canto, Damjan Cicin-Sain, Astrid Hoermann, Hilde
  Janssens, Julio~R Banga, and Johannes Jaeger.
\newblock Reverse-engineering post-transcriptional regulation of gap genes in
  drosophila melanogaster.
\newblock \emph{PLoS computational biology}, 9\penalty0 (10):\penalty0
  e1003281, 2013.

\bibitem[Duane et~al.(1987)Duane, Kennedy, Pendleton, and Roweth]{duane87}
Simon Duane, Anthony~D Kennedy, Brian~J Pendleton, and Duncan Roweth.
\newblock Hybrid monte carlo.
\newblock \emph{Physics letters B}, 195\penalty0 (2):\penalty0 216--222, 1987.

\bibitem[Girolami and Calderhead(2011)]{girolami11}
Mark Girolami and Ben Calderhead.
\newblock Riemann manifold langevin and hamiltonian monte carlo methods.
\newblock \emph{Journal of the Royal Statistical Society: Series B (Statistical
  Methodology)}, 73\penalty0 (2):\penalty0 123--214, 2011.

\bibitem[Graepel(2003)]{graepel03}
Thore Graepel.
\newblock Solving noisy linear operator equations by gaussian processes:
  Application to ordinary and partial differential equations.
\newblock In \emph{ICML}, pages 234--241, 2003.

\bibitem[Jaeger(2011)]{jaeger11}
J.~Jaeger.
\newblock {{T}he gap gene network}.
\newblock \emph{Cell. Mol. Life Sci.}, 68\penalty0 (2):\penalty0 243--274, Jan
  2011.

\bibitem[Jaeger et~al.(2004)Jaeger, Surkova, Blagov, Janssens, Kosman, Kozlov,
  Manu, Myasnikova, Vanario-Alonso, Samsonova, Sharp, and Reinitz]{jaeger04}
J.~Jaeger, S~Surkova, M~Blagov, H~Janssens, D~Kosman, KN~Kozlov, Manu,
  E~Myasnikova, CE~Vanario-Alonso, M~Samsonova, DH~Sharp, and J~Reinitz.
\newblock Dynamic control of positional information in the early
  \emph{Drosophila} embryo.
\newblock \emph{Nature}, 430\penalty0 (6997):\penalty0 368--371, 2004.

\bibitem[Jaeger et~al.(2007)Jaeger, Sharp, and Reinitz]{jaeger07}
J.~Jaeger, D.~H. Sharp, and J.~Reinitz.
\newblock {{K}nown maternal gradients are not sufficient for the establishment
  of gap domains in \emph{{D}rosophila melanogaster}}.
\newblock \emph{Mech. Dev.}, 124\penalty0 (2):\penalty0 108--128, Feb 2007.

\bibitem[Lawrence et~al.(2006)Lawrence, Sanguinetti, and Rattray]{lawrence06}
Neil Lawrence, Guido Sanguinetti, and Magnus Rattray.
\newblock Modelling transcriptional regulation using gaussian processes.
\newblock 2006.

\bibitem[Leimkuhler and Reich(2004)]{leimkuhler04}
Benedict Leimkuhler and Sebastian Reich.
\newblock \emph{Simulating hamiltonian dynamics}, volume~14.
\newblock Cambridge University Press, 2004.

\bibitem[Liu(2008)]{liu08}
Jun~S Liu.
\newblock \emph{Monte Carlo strategies in scientific computing}.
\newblock springer, 2008.

\bibitem[Murray-Smith and Pearlmutter(2005)]{murray05}
Roderick Murray-Smith and Barak~A Pearlmutter.
\newblock Transformations of gaussian process priors.
\newblock In \emph{Deterministic and Statistical Methods in Machine Learning},
  pages 110--123. Springer, 2005.

\bibitem[Neal(2011)]{neal11}
Radford Neal.
\newblock Mcmc using hamiltonian dynamics.
\newblock \emph{Handbook of Markov Chain Monte Carlo}, 2, 2011.

\bibitem[Neal(1996)]{neal96}
Radford~M Neal.
\newblock Sampling from multimodal distributions using tempered transitions.
\newblock \emph{Statistics and computing}, 6\penalty0 (4):\penalty0 353--366,
  1996.

\bibitem[Nusslein-Volhard and Wieschaus(1980)]{nuss80}
C.~Nusslein-Volhard and E.~Wieschaus.
\newblock {{M}utations affecting segment number and polarity in
  \emph{{D}rosophila}}.
\newblock \emph{Nature}, 287\penalty0 (5785):\penalty0 795--801, Oct 1980.

\bibitem[O'Hagan and Kingman(1978)]{o78}
Anthony O'Hagan and JFC Kingman.
\newblock Curve fitting and optimal design for prediction.
\newblock \emph{Journal of the Royal Statistical Society. Series B
  (Methodological)}, pages 1--42, 1978.

\bibitem[Papoulis and Pillai(2002)]{papoulis02}
Athanasios Papoulis and S~Unnikrishna Pillai.
\newblock \emph{Probability, random variables, and stochastic processes}.
\newblock Tata McGraw-Hill Education, 2002.

\bibitem[Pisarev et~al.(2009)Pisarev, Poustelnikova, Samsonova, and
  Reinitz]{flyex2}
A.~Pisarev, E.~Poustelnikova, M.~Samsonova, and J.~Reinitz.
\newblock {{F}ly{E}x, the quantitative atlas on segmentation gene expression at
  cellular resolution}.
\newblock \emph{Nucleic Acids Res.}, 37\penalty0 (Database issue):\penalty0
  D560--566, Jan 2009.

\bibitem[Poustelnikova et~al.(2004)Poustelnikova, Pisarev, Blagov, Samsonova,
  and Reinitz]{flyex1}
E.~Poustelnikova, A.~Pisarev, M.~Blagov, M.~Samsonova, and J.~Reinitz.
\newblock {{A} database for management of gene expression data in situ}.
\newblock \emph{Bioinformatics}, 20\penalty0 (14):\penalty0 2212--2221, Sep
  2004.

\bibitem[Rasmussen(2006)]{rasmussen06}
Carl~Edward Rasmussen.
\newblock Gaussian processes for machine learning.
\newblock 2006.

\bibitem[Robert and Casella(2004)]{robert04}
Christian~P Robert and George Casella.
\newblock \emph{Monte Carlo statistical methods}, volume 319.
\newblock Citeseer, 2004.

\bibitem[Rogers et~al.(2006)Rogers, Khanin, and Girolami]{rogers06}
Simon Rogers, Raya Khanin, and Mark Girolami.
\newblock Model based identification of transcription factor activity from
  microarray data.
\newblock \emph{Probabilistic Modeling and Machine Learning in Structural and
  Systems Biology, Tuusula, Finland}, 2006.

\bibitem[S{\"a}rkk{\"a}(2011)]{sarkka11}
Simo S{\"a}rkk{\"a}.
\newblock Linear operators and stochastic partial differential equations in
  gaussian process regression.
\newblock In \emph{Artificial Neural Networks and Machine Learning--ICANN
  2011}, pages 151--158. Springer, 2011.

\bibitem[Surkova et~al.(2009)Surkova, Spirov, Gursky, Janssens, Kim, Radulescu,
  Vanario-Alonso, Sharp, Samsonova, Reinitz, et~al.]{surk09}
Svetlana Surkova, Alexander~V Spirov, Vitaly~V Gursky, Hilde Janssens, Ah-Ram
  Kim, Ovidiu Radulescu, Carlos~E Vanario-Alonso, David~H Sharp, Maria
  Samsonova, John Reinitz, et~al.
\newblock Canalization of gene expression and domain shifts in the drosophila
  blastoderm by dynamical attractors.
\newblock \emph{PLoS computational biology}, 5\penalty0 (3):\penalty0 e1000303,
  2009.

\end{thebibliography}
\bibliographystyle{plainnat}

\end{document}